\documentclass{article}

\usepackage{PRIMEarxiv}

\usepackage[utf8]{inputenc} 
\usepackage[T1]{fontenc}    
\usepackage{url}            
\usepackage{booktabs}       
\usepackage{amsfonts}       
\usepackage{nicefrac}       
\usepackage{microtype}      
\usepackage{fancyhdr}       

\usepackage{times}
\usepackage{epsfig}
\usepackage{graphicx}
\usepackage{amsmath}
\usepackage{amssymb}
\usepackage{caption}
\usepackage{subcaption}
\usepackage{multirow}
\usepackage{xcolor}
\usepackage{balance}

\usepackage[pagebackref=true,breaklinks=true,letterpaper=true,colorlinks,bookmarks=false]{hyperref} 

\begin{document}

\title{Implicit neural representation for change detection}

\author{Peter Naylor\\
RIKEN AIP\\
Kyoto, Japan\\
{\tt\small peter.naylor@riken.jp}
\and
Diego Di Carlo\\
RIKEN AIP\\
Kyoto, Japan\\
{\tt\small diego.dicarlo@riken.jp}
\and
Arianna Traviglia\\
Istituto Italiano di Tecnologia\\
Venice, Italy\\
{\tt\small arianna.traviglia@iit.it}
\and
Makoto Yamada\\
OIST\\
Okinawa, Japan\\
{\tt\small makoto.yamada@oist.jp}
\and
Marco Fiorucci\\
Istituto Italiano di Tecnologia\\
Venice, Italy\\
{\tt\small marco.fiorucci@iit.it}
}

\fancyhead[LO]{}

\maketitle

%
%

\newcommand{\NFacr}{INR}
\newcommand{\NFfull}{Implcit Neural Represenation}

\begin{abstract}
Identifying changes in a pair of 3D aerial LiDAR point clouds, obtained during two distinct time periods over the same geographic region presents a significant challenge due to the disparities in spatial coverage and the presence of noise in the acquisition system.
The most commonly used approaches to detecting changes in point clouds are based on supervised methods which necessitate extensive labelled data often unavailable in real-world applications.  
To address these issues, we propose an unsupervised approach that comprises two components: \NFfull{} (\NFacr) for continuous shape reconstruction and a Gaussian Mixture Model for categorising changes.
\NFacr{} offers a grid-agnostic representation for encoding bi-temporal point clouds, with unmatched spatial support that can be regularised to enhance high-frequency details and reduce noise.
The reconstructions at each timestamp are compared at arbitrary spatial scales, leading to a significant increase in detection capabilities.
We apply our method to a benchmark dataset comprising simulated LiDAR point clouds for urban sprawling. This dataset encompasses diverse challenging scenarios, varying in resolutions, input modalities and noise levels. This enables a comprehensive multi-scenario evaluation, comparing our method with the current state-of-the-art approach. We outperform the previous methods by a margin of $10\%$ in the intersection over union metric.
In addition, we put our techniques to practical use by applying them in a real-world scenario  to identify instances of illicit excavation of archaeological sites and validate our results by comparing them  with findings from field experts.

\end{abstract}
 
\section{Introduction}

\begin{figure*}[!ht]
    \centering
    \begin{subfigure}[b]{0.29\textwidth}
          \centering
          \includegraphics[width=\textwidth]{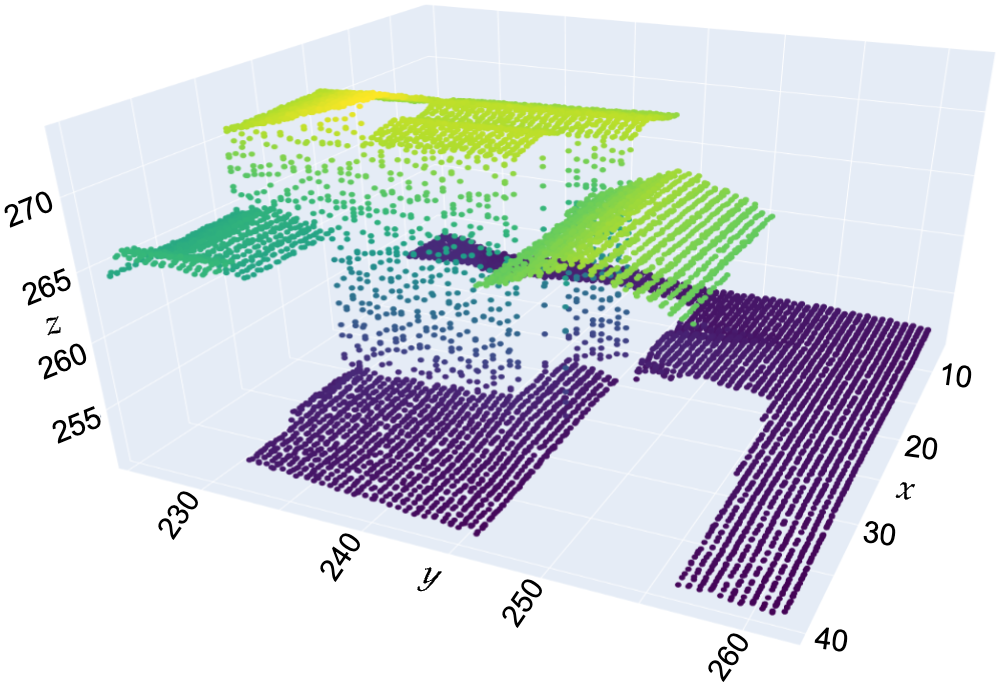}
          \caption{Point cloud at $t=0$ $(pc_0)$}
          \label{fig:pc0}
    \end{subfigure}
    \hfill
    \begin{subfigure}[b]{0.33\textwidth}
          \centering
          \includegraphics[width=\textwidth]{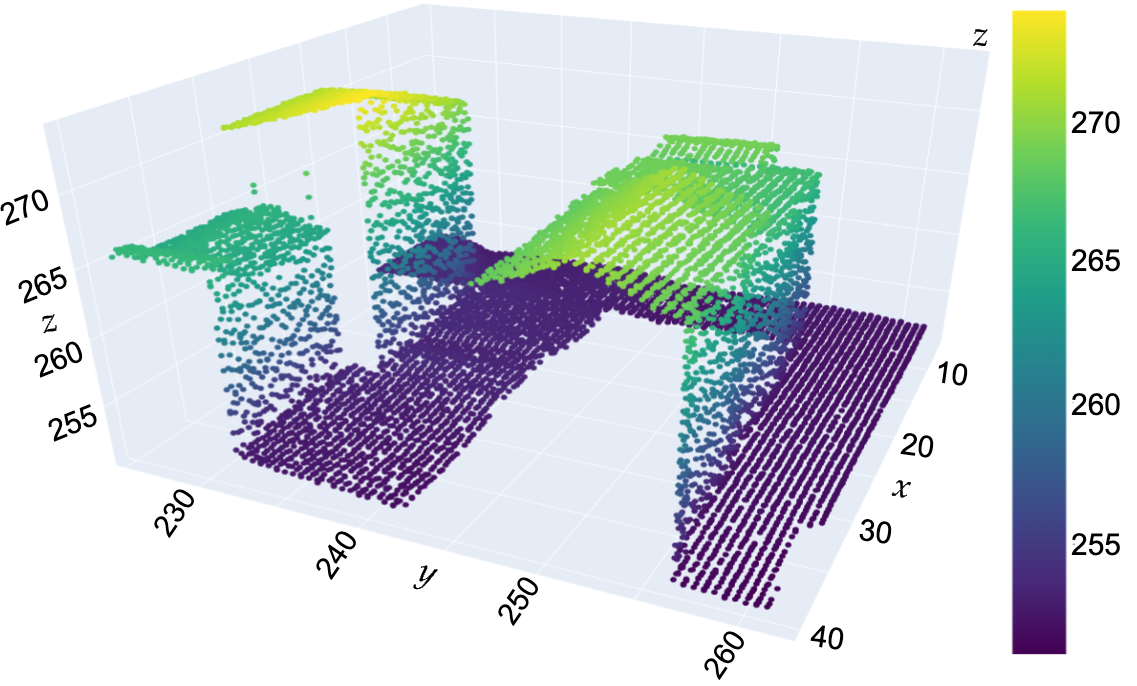}
          \caption{Point cloud at $t=1$ $(pc_1)$}
          \label{fig:pc1}
    \end{subfigure}
    \hfill
    \begin{subfigure}[b]{0.29\textwidth}
          \centering
          \includegraphics[width=\textwidth]{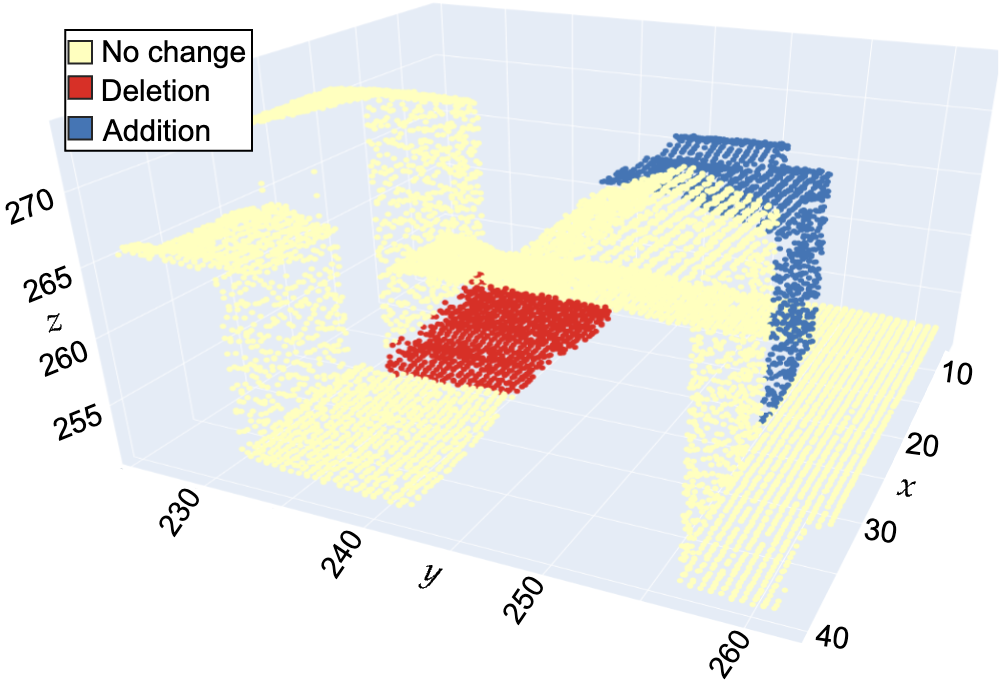}
          \caption{$pc_1$ with annotated change}
          \label{fig:pc1change}
    \end{subfigure}
       \caption{Simulated airborne LiDAR data for change detection of a clipped test data area.}
       \label{fig:lidar}
 \end{figure*}
In contemporary times, we observe the Earth through various sensors at unprecedented spatial and temporal resolutions. 
One of the most popular Earth Observation technologies is LiDAR: by using laser light to measure distances, it creates detailed three-dimensional maps or point cloud representations of the Earth's surface and objects, see Fig.~\ref{fig:lidar}.
LiDAR has been applied to autonomous driving \cite{Li2020}, robotics \cite{nubert2022}, digital terrain mapping \cite{vosselman2010airborne}, city planning \cite{park2019creating}, urban sprawling \cite{de2021change, de2023dc3dcd, Kharroubi2022}, and cultural heritage \cite{fiorucci_2022, sech2023, verschoof_2022_thesis}. 
The surge in popularity can be attributed to three main factors.
Firstly, LiDAR data exhibit a remarkable level of precision, with spatial resolutions typically falling below 1 m in most applications (though they can be larger based on the distance to the scanned scene or specific detector characteristics), effectively capturing the intricate details of a 3D environment.
Secondly, LiDAR acquisition systems remain unaffected by varying lighting conditions.
Lastly, LiDAR possesses the capability to map terrain and unveil structures concealed by vegetation canopies  \cite{canuto2018ancient, chan2021estimating}. 

The process of comparing two or more slightly co-registered EO data to identify and analyse discrepancies that have emerged between them is called Change Detection (CD). 
In this work, we focus on the application of CD to urban sprawl, shown in Fig.~\ref{fig:lidar}, and on the identification of illicit excavations of archaeological sites (looting) by detecting relevant changes in height.
Urban sprawl monitoring, which entails the identification of recently erected and demolished structures within multi-temporal LiDAR point cloud datasets, has been identified as a method to assist landscape and city managers in promoting sustainable development \cite{de2021change}. 
\begin{figure}[!ht]
    \centering
    \includegraphics[trim={0 80 150 0},clip,width=0.7\linewidth]{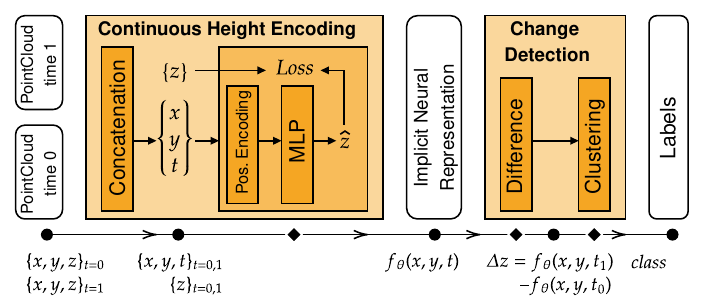}
    \caption{\small Proposed unsupervised CD of PCs based on implicit neural representation and clustering.}
    \vspace{-0.5cm}
    \label{fig:pipeline}
\end{figure}
The identification of illicit excavations of archaeological sites holds paramount significance due to the potential for looting to cause damage, displacement, or irrevocable loss of priceless archaeological artefacts.

The majority of existing techniques handle data that are defined on discrete regular grids (e.g., images) with supervised learning frameworks.
Over the past decade, Deep Neural Networks (DNN) have risen to prominence as the state-of-the-art solution for CD on LiDAR data \cite{de2021benchmarking, Kharroubi2022, Zhang2019}.
However, these approaches require a preprocessing step to cleanse and project a raw 3D LiDAR point cloud onto a consistent 2D regular grid with predefined spatial resolution.
The prevailing  projection techniques encompass the use of the Digital Elevation Model (DEM) \cite{OKYAY2019} and Digital Surface Model (DSM) \cite{Erdogan2019, Lyu2020, Zhang2019}.
Consequently, the accuracy of these models hinges on the accuracy of the projection, and memory complexity increases proportionately with the desired resolution, limiting their application to large point clouds (PCs).
In contrast, our investigation delves into the application of CD directly onto the off-grid, raw 3D LiDAR PCs in an unsupervised manner, which suits better real-world applications.




In summary, our approach involves two main steps, illustrated in Fig.~\ref{fig:pipeline}.
In the first step, we use \NFacr \cite{xie2022neural} to encode the bi-temporal PC as a continuous function of both time and space. This function is estimated blindly from data using the total variation (TV) norm \cite{bleakley2011group} to enforce discontinuities along the time dimension to model sharp temporal changes and increase robustness to noisy measurements.
In the second step, we categorise the change in altitude given by decoding the surface at both timestamps at an arbitrary resolution.
To validate the proposed approach, we consider two EO datasets.
The first is an open simulated airborne LiDAR dataset comprising 15 distinct PCs. The goal here is to identify both newly built and demolished buildings. 
The second dataset consists of a pair of PCs over the Kulen, a region of Cambodia featuring numerous archaeological structures lying under the forest where instances of illegal excavations have been recorded.
Our goal is to precisely pinpoint locations where looting has taken place and to record the shape and attributes of the looting pits.

\textit{Contributions.} This paper tackles the lack of extensively labelled 3D PCs for CD in real-world applications by introducing an effective unsupervised pipeline. Specifically:
\begin{itemize}
   \itemsep0em 
   \item Using a single and scalable Implicit Neural Representation to encode the height of two geo-referenced 3D point clouds as a continuous quantity.
   \item Denoising  and regularising the encoded surface with total variation norm over the time dimension.
   \item Proposing an end-to-end unsupervised change detection pipeline that leverages on automatic hyper-parameter tuning and a simple clustering algorithm. 
\end{itemize}

\section{Related works}

The off-grid and noisy nature of PCs collected by a LiDAR system detrimentally impacts the performances of CD methods.
Moreover, weather conditions and remote sensor trajectory may differ for two measurements at different timestamps, leading to spatial unmatching support.
Despite this, supervised methods for CD can achieve great performances even in challenging noisy data. 
However, their usage is still limited to a few real-world applications due to the prohibitive cost of collecting and labelling datasets.

For these reasons, unsupervised methods represent an attractive alternative. 
Such methods can be broadly classified into three categories~\cite{de2021change}: those based on distance computation \cite{girardeau2005change,shirowzhan2019comparative}, those based on optimal transport \cite{courty2016optimal, fiorucci2023optimal} and those adapting PC DNN for unsupervised learning \cite{de2023deep,de2023dc3dcd}.
Distance-based methods, such as C2C \cite{girardeau2005change}, and M3C2 \cite{lague2013accurate,shirowzhan2019comparative} divide the PC via octrees, estimate the surface normal and orientation to calculate pair-wise euclidean distances.
Alternatively, optimal transport-based methods estimate a distance based on the projection matrix of the first PC onto the second.
In both cases, the actual changes are then classified via empirical thresholding, or the OTSU method \cite{otsu1979threshold}.
These methods have been developed and applied to the only available airborne LiDAR dataset for building CD \cite{de2021change}. 
However, datasets usually contain millions of points because they are acquired with high-spatial resolution acquisition systems \cite{Schwarz_2010}. 
The previous methods do not scale well with the data size and have to subdivide the PC for analysis.
The adaptation of PC DNN for unsupervised learning are very recent and were absent at the initial time of writing (they have yet to be peer-reviewed) \cite{de2023deep,de2023dc3dcd}.
PC DNN use Kernel Points \cite{thomas2019kpconv} to compare co-referenced subsets of the two PCs.
SSL-DCVA uses intermediate representation of the cylinders in the DNN to build an estimator for change \cite{de2023deep} inspired by DCVA \cite{saha2019unsupervised}.
DC3DCD inspires itself from DeepCluster \cite{caron2018deep} with pseudo-labels for training.
Based on recent reviews \cite{stilla2023change,xiao20233d}, the aforementioned methods are the only unsupervised methods and fully automatic state-of-the-art approaches for PC CD.

Other methods for unsupervised CD do not use raw 3D LiDAR data directly.
Instead, they used 2D images obtained by projecting the 3D PCs on a 2D regular grid, e.g., for DEM and DSM \cite{Stular_2012}. 
Due to the grid's regularity, ordering and consistency, these projected 2D images are ideally suited for convolutional operations. 
Hence, convolutional neural network-based architectures can be applied to the 2D digital models for building CD \cite{Shi2020_RS, Zhang2019, Zhang2018ChangeDB}.
However, these projections lead to precision loss of the LiDAR data as the height measurements are interpolated to output the DSM or the DEM \cite{de2021change}. 
In contrast, we use \NFacr{} to model the surface and obtain an intermediate 2D images at any desired resolution of the surfaces height.

Novel DNN models, called \NFacr s~\cite{xie2022neural}, have tackled off-grid PC analysis \cite{Kharroubi2022, qi2017pointnet} and 3D surface reconstruction~\cite{huang2022surface, sitzmann2019siren, tancik2020fourier} in both supervided and unsupervised way.
The \NFacr s are \textit{coordinate}-based continuous deep learning models that map coordinates to a target value, called field (e.g. from pixel position to colour in an image).
This function can be estimated by fitting observation and inducing desired properties by application-based regularisation terms. The major property of these parametric models is their capability to natively interpolate the target field.

A common limitation of \NFacr{} is their poor capability of capturing high-frequency details of the surface, referred to as spectral bias.
To address this, positional encoding and Random Fourier Features (RFF)~\cite{rahimi2007random,tancik2020fourier} of the input coordinates have become standard practice.
This allowed to apply \NFacr{} to a new plethora of applications and methodologies, such as 3D shape reconstruction from sparse images, animation of human bodies and faces, as well as video coding~\cite{xie2022neural}.
Moreover, RFF have improved Physics-Informerd Neural Networks (PINNs)~\cite{raissi2019physics}, which are coordinate-based DNNs trained to fit observations while solving partial differential equation evaluated with network's backpropagation algorithm~\cite{di2022post, karniadakis2021physics}.

An alternative approach to surpass the spectral bias is SIREN~\cite{sitzmann2019siren}, where the standard nonlinearities are replaced with periodic sine functions. 
Such architectures show a very low reconstruction error, both in fitting the target field and its gradient with respect to the input. 
SIRENs have become standard backbone networks for both \NFacr{} and PINNs~\cite{xie2022neural}.
Despite their success, SIRENs suffer from difficulties in training as they are prone to overfit or being stuck in local minima, for which careful parameter tuning is required.
These issues have been addressed in subsequent works~\cite{fathony2021multiplicative,lindell2022bacon}, where novel architectures have been proposed claiming easier training and faster convergence. 
Nevertheless, SIREN's performance depends on the applications. 
As reported in this work, SIREN doesn't consistently outperform RFF-based architecture.

Most of the \NFacr-based models for 3D reconstruction recover the object shape by minimising the signed distance between any given point and the closest surface~\cite{park2019deepsdf}.
Then, the associated \NFacr{} is a function of the three spatial coordinates.
A positive sign implies the point lies outside the object and vice versa.
The shape contours are then found by checking the $0$ iso-line of the learned function.
Neural Unsigned Distance Fields \cite{chibane2020neural} remove the sign from the distance function and encode continuous locations for a stronger regularisation, leading to better reconstruction.
Nonetheless, these methods require knowledge of the normals vectors of each point within the PC~\cite{huang2022surface}. 
While the rudimentary normal estimation of LiDAR data might be achieved using conventional geometrical methods, our preliminary investigations have demonstrated that errors stemming from such preprocessing steps significantly undermine the performances.
As elucidated in Section~\ref{sec:method}, the phenomenon of urban sprawl can be reasonably approximated through a continuous function across a 2D spatial domain, while the exploration of 3D aspects will be deferred to forthcoming research endeavors.

\section{Method} \label{sec:method}
We want to detect changes in two geo-referenced LiDAR PCs with unsupervised methods.
A PC at time $t$ will be denoted by $\mathcal{X}_t \subset \mathbb{R}^3$ where each element is a 3D coordinate, i.e. $(x, y, z) \in \mathcal{X}_t$.
We will denote by $t_0=0$ and $t_1=1$ the two timestamps for which we wish to detect change.
If the support of $\mathcal{X}_0$ and $\mathcal{X}_1$ match, we naturally define the addition of an element by a positive difference, i.e. the 2D point $(x, y)$ is of the label "Addition" if the associated altitudes: $z_1 - z_0 > \alpha$, where $z_t$ corresponds to the altitude at time $t$ and with $\alpha$ a fixed scalar.
Similarly, we can define the "Deletion" class by a negative difference.

In general, the above operations are not directly applicable to LiDAR PCs as the supports do not match due to different acquisition conditions.
To fix the support, some have used projection methods \cite{de2021change} and optimal transport \cite{fiorucci2023optimal}.
In the current paper, we fix the support matching by estimating a surface from the PC, allowing us to interpolate and query any spatial point.
In other words, we reconstruct the surface at a given time point.
We can then compute the difference to find additions or deletions.
It is important to note that we are estimating a function that maps a position to an altitude, which is different from the actual PC. 
Some points will share the same $(x,y)$ but have different altitudes due to the inclination of the LiDAR emitter and the verticality of elements in the maps, especially with buildings. 

\subsection{Regression model}
We denote by $f_{\theta}$ a DNN model with learnable parameters $\theta$.
Given an input vector $\mathbf{v}$, we estimate the density by $f_{\theta}$ with inputs in $V$ and values in $\mathbb{R}$.
As a baseline, we can independently reconstruct the first and second PC with $V=\mathbb{R}^2$.
We learn two functions and name this model (D).
The formulae for CD with the baseline method is:
\begin{equation}
 \Delta z(x, y) = (f_{\theta_1} - f_{\theta_0})(x,y).
\label{eq:loss2}
\end{equation}
Here $\theta_t$ corresponds to the set of DNN parameters to reconstruct the surface at time $t$ that are optimised by minimising the mean squared error (MSE) between the estimated and observed altitude. 

A more compact and efficient representation is possible where we modify the input $\mathbf{v}\in\mathbb{R}^3$ of the network to incorporate time. 
In particular, we learn a single model (S) with parameters $\theta$, and the detection formula is modified to:
\begin{equation}
\Delta z(x, y) = f_{\theta}(x,y,t_1) - f_{\theta}(x,y, t_0) \label{eq:loss1}
\end{equation}
\textit{A priori}, it is not evident to know in advance which method is better suited for a given dataset.
In some configurations of the simulated datasets, $\mathcal{X}_0$ and $\mathcal{X}_1$ are not drawn from the same distribution, which could potentially be harmful to the single model.
Therefore we investigate the benefits of a single model as opposed to two.
In the following sections, we will describe in more detail the different regularisations and model specificities we apply to improve reconstruction.

\subsection{Random Fourier Features}

As common in INR models, we map the input $\mathbf{v}$ to a higher dimensional space with RFF. It has been shown that this projection is crucial for estimating high frequencies for a better reconstruction \cite{tancik2020fourier}.
For a fixed model, RFF is defined as follows:
\begin{align*}
& B \in \mathbb{R}^{M\times 3}, \ \forall (i,j), B_{ij} \sim \mathcal{N}(0, \sigma)   \\
& \gamma_B(\mathbf{v}) = [\cos(2\pi B\mathbf{v}),~\sin(2\pi B\mathbf{v})] 
\end{align*}
The size of $B$ depends on the size of $\mathbf{v}$, and here we supposed $\mathbf{v} \in \mathbb{R}^3$.
The mapping size $M$ and the scale $\sigma$ are two hyper-parameters that need tuning.

\subsection{Network architecture}

We propose several network architectures comprising MLP layers, different activations, and skip layers in Table~\ref{tab:model-recap} of the Appendix. 
We show in Fig.~\ref{fig:model} a particular model that we name `skip-ten-only'.
In particular, the network size will depend on the data complexity.
We use skip layers allowing a better gradient flow \cite{he2016deep}.
The activation functions will be ReLu or hyperbolic tangent (tanh), as tanh allows the resulting  $f_{\theta}$ to be $\mathcal{C}^\infty$, we do not use any activation layers for the last layer.
We use a similar architecture to the one presented in Fig.~\ref{fig:model} for the SIREN methods. 
We allow the final model to fine-tune the architecture for both methods by making it a hyperparameter.

\subsection{Total variation norm (TVN)}
The total variation norm, defined as $\sum |u_{i+1} - u_i|$ is a standard regularisation scheme for sequential like data \cite{bleakley2011group}.
It is also helpful to smooth spatial patterns, as a sudden altitude change should be penalised.
Such a scheme has been generalised to a continuous version by enforcing that the resulting gradient of the function $f_{\theta}$ be sparse over the spatial coordinates ~\cite{raissi2019physics}.
It is possible to add to the loss function the regularisation term $\mathcal{R}_{TV} = \lvert \frac{\partial f_{\theta}}{\partial x} \rvert + \lvert \frac{\partial f_{\theta}}{\partial y} \rvert$.

\subsection{Time difference (TD)}
Similarly to the discrete total variation norm, we can enforce that the change over time be sparse, which indicates to the DNN that most points do not change over time, but we will allow some to change. 
To enforce such a constraint, we add the following regularisation to the loss function: $\mathcal{R}_{TD} = \lvert f_{\theta}(x, y, t_1) - f_{\theta}(x, y, t_0) \rvert$.
This regularisation is only possible when the input $\mathbf{v}$ contains time, and therefore it cannot be applied where we reconstruct the surface for two models (\ref{eq:loss2}). 
This regularisation over time is very similar to the total variation norm over the temporal domain.
We stress that shape reconstruction does not need this regularisation term, and this is only a CD regulariser.
In particular, adding this term to the loss allows us to fuse the information from both PCs more efficiently and enables the PCs to benefit from each other mutually.

\subsection{SIREN}

From an architectural point of view, the SIREN network~\cite{sitzmann2019siren} is a simple modification of an MLP where standard activation functions, e.g. ReLU and tanh, are continuous sine functions.
This substitution enables the modelling of a continuous complicated signal without the need for explicit upsampling in various domains. The SIREN network can then be described by
\begin{equation}
\begin{split}
    f({\bf x}) &= \mathbf{W}_n (g_{n-1} \circ g_{n-2} \circ \ldots \circ g_0{\mathbf{x}}) + \mathbf{h}_n, \\
    & \text{where} \quad g_i(\mathbf{h}_i) = \sin(\mathbf{W}_i \mathbf{h}_i + \mathbf{h}_i).
\end{split}
\end{equation}
SIREN operates as a composition of sinusoidal transformations recalling the principle behind RFF. 
In fact, it has been shown that positional encoding with RFF is equivalent to periodic nonlinearities with one hidden layer as the first DNN  layer~\cite{benbarka2022seeing}.
Good initialisation of the weights is critical for their successful training.
To avoid saturation of the sine activations, a scalar hyperparameter $\sigma$ is introduced to scale the layers' weights, like for RFF.

\subsection{Unsupervised labelling of $\Delta z$}

Once the surface reconstruction is performed, we have access to $\Delta z$ given by equation (\ref{eq:loss2}) or (\ref{eq:loss1}).
The OTSU threshold method has been used to separate binary sources \cite{naylor2018segmentation, otsu1979threshold} for CD.
However, in our case, we have three sources to distinguish and use a Gaussian Mixture Model (GMM) with three components \cite{mclachlan1988mixture}. 
We show in Fig.~\ref{fig:diffz} the application of GMM to $\Delta z$ for a small clipped sample.
The success  of the GMM depends on the distribution of $\Delta z$. 
GMM will divide the distribution into three, regardless of the shape of the distribution and lead to a random score.

\section{Experimental setting}

\subsection{Simulated dataset}

We use the publicly simulated airborne LiDAR dataset for CD: Urb3DCD \cite{de2021change}.
Even if this dataset is simulated, it mimics true data with different noise levels and sensors used in practice.
Five simulation configurations are given: 1 - low resolution -- low noise, 2 - high resolution -- low noise, 3 - low resolution -- high noise, 4 - photogrammetry and 5 - multi-sensor.
The photogrammetry setting is low resolution, high noise and tight scan angle for each timestamp. They mimic satellite acquisition.
The multi-sensor simulation is characterised by $pc_0$ and $pc_1$ having different resolutions and noise levels. 
$pc_0$ is low resolution and high noise, whereas $pc_1$ is high resolution and low noise. 
In this situation $\# pc_1 >> \# pc_0$ which is different to other subsets where $\# pc_1 \approx \# pc_0$.
$\#$ denotes the cardinality of the set.
Each configuration is divided into training and testing datasets.
We will only apply the method to the testing sets as the methods used are unsupervised.
For each testing configuration, three simulated datasets exist where the ground truth is different for each.
The testing set comprises three different geographical areas of the city of Lyon in France \cite{de2021change}.
Only the second PC $pc_1$ is annotated with additions and deletion changes and will be used to evaluate the methods.
In Fig. \ref{fig:pc1change}, we overlay $pc_1$ with the annotation.

This dataset was updated (Urb3DCD-v2) to include vegetation changes and mobile objects in two simulation settings, low density and multi-sensor LiDAR acquisition.
This version is used to compare the DC3DCD model \cite{de2023dc3dcd}. 

 
\subsection{Metrics}
The minimised MSE used to cross-validate training will not be used for evaluation.
Due to the noise level, a good performance on this metric will not imply a good reconstruction.
Indeed an MSE of $ 0 $ implies that the model perfectly reconstructs the data and the noise.

\textbf{Intersection over union} uses predicted and true labels: $IoU(P, G) = \frac{P \cap G}{P \cup G}$.
This metric is very sensitive to small changes, especially when the ground truth is small, like in our situation. 
The $IoU$ will be measured after applying the GMM to $\Delta z$.
In particular, a low score could mean that the GMM is unfit for converting the differences into labels or that the surface reconstruction failed.

\textbf{Average AUC} will be computed to highlight good-performing methods irrespective of the GMM results.
We compute the standard AUC over three settings: addition vs no addition, deletion vs no deletion and change vs no change.
In the last setting, we use $\lvert \Delta z \rvert$.

\subsection{Training procedure for surface reconstruction}
In detail, we will describe how we train our network $f_\theta$ given a PC.
When a single network is used, $pc_0$ and $pc_1$ are concatenated, and the input dimension is three.
With no loss of generality, we will consider that we have a single PC $pc$ of dimension two or three.
We normalise the PC to be in $[-1;1]$ on each axis which is a requirement for the methods RFF and SIREN \cite{sitzmann2019siren}.
We randomly split the dataset in two where $80\%$ is retained for training and the other $20\%$ is used for validating the surface reconstruction.
We minimise and backpropagate through the training loss and evaluate the MSE performance on the validation.
We use Optuna \cite{akiba2019optuna} to find the best set of hyperparameters via bayesian optimisation that minimises the validation MSE.
The tuned hyper-parameters are the model architecture, the learning rate, the batch size, the scale of the gaussian mapping, the scalars associated with the regularisation terms $\lambda_{TD}$, and $\lambda_{TV}$.
When we use the SIREN model, we also optimise the scale size by multiplying the signal in the sinusoidal activation function, the number of layers, and the number of hidden units for each layer.
We use the optimisation method Adam \cite{kingma2014adam} wrapped with the Layer-wise Adaptive Rate Scaling (LARS) \cite{torchlars} that enables the use of an enormous batch size that for us is essential to carry out our experiments in a reasonable time.
To speed up computation, we set the number of epochs to 50, use learning rate decay and early stopping.
To compute the TV norm, we sample random elements from $pc$ that we corrupt with noise and backpropagate their prediction to the input \cite{raissi2019physics}.

\section{Results  and discussion}
\subsection{Feature mapping} \label{sec:featuremap}
\begin{figure}[t]
    \centering
    \includegraphics[width=0.6\linewidth]{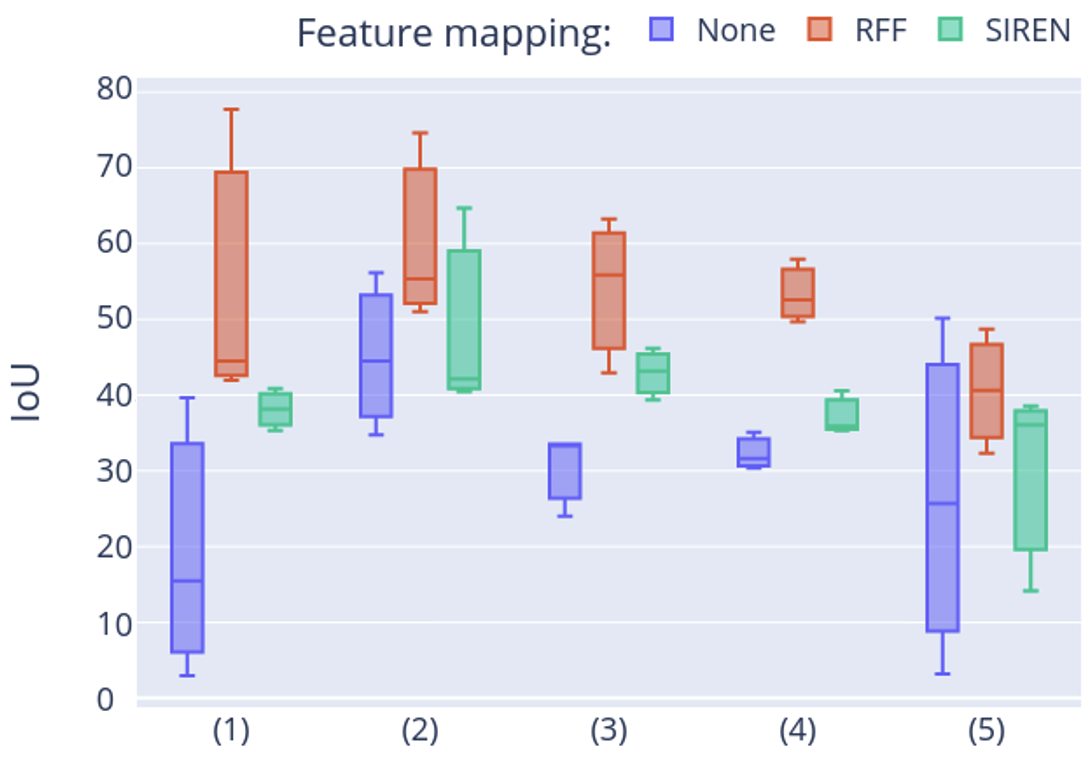}
    \caption{$IoU$ results (in $\%$) for different feature mapping methods for every LiDAR airborne simulated dataset.}\vspace{-0.5cm}
    \label{fig:results_feature_mapping}
 \end{figure}
\begin{figure*}[t]
    \centering
    \includegraphics[width=1.\linewidth]{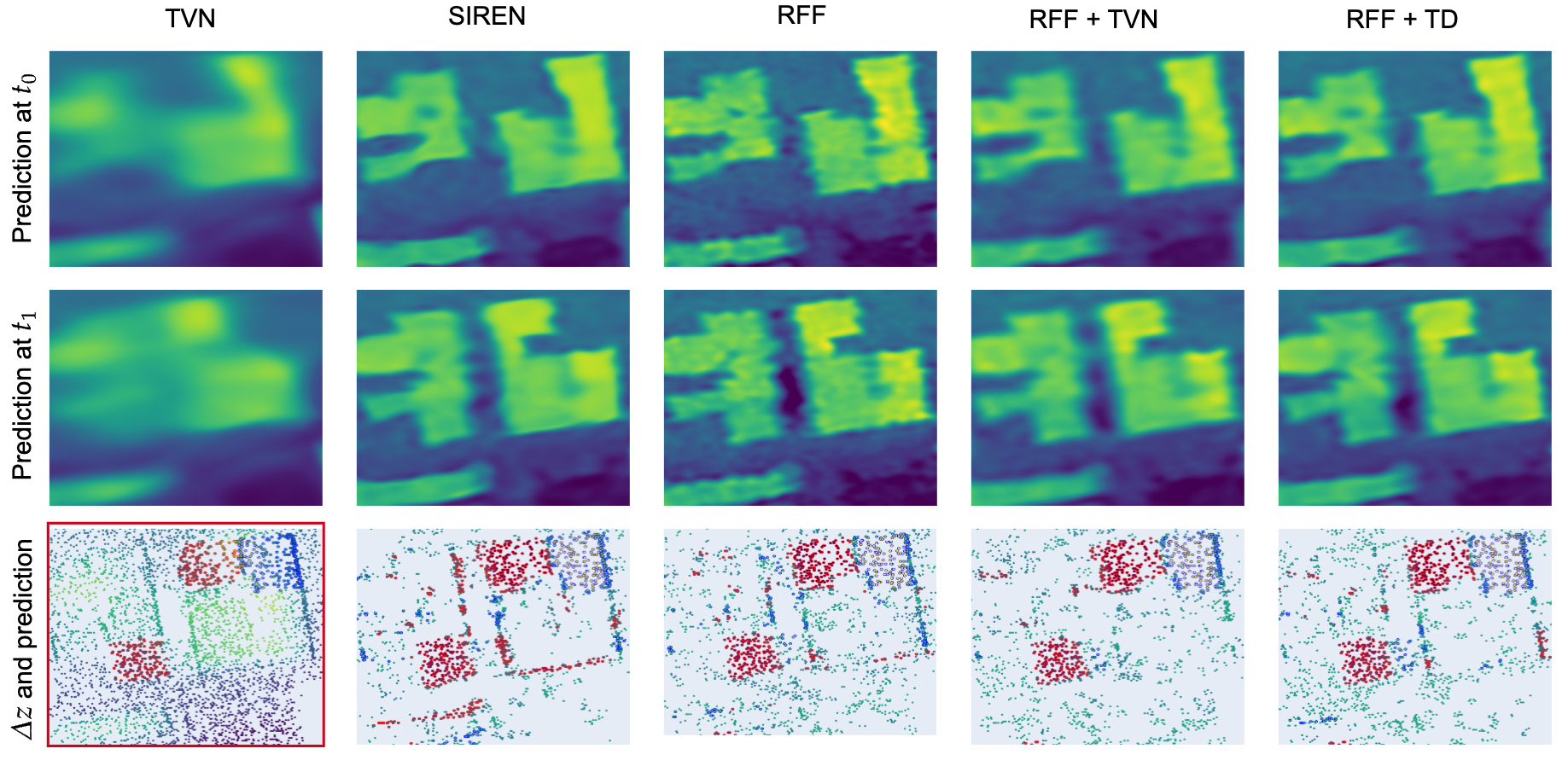}
    \caption{Visualisation of a crop where in each column we show a different method comprising a single DNN trained on dataset (3). In the two first rows, we reconstruct the surface along a regular grid for timestamp $t_0$ and $t_1$. In the third row, we show the difference $\Delta z$ on the support of $\mathcal{X}_1$ with it's predicted labels from the GMM, we filter out points where $|\Delta z| < 2 m$. Each column shows a different method. In the final row and in the first column we show the true cloud point overlaid with the ground truth. To compare fairely, the color map ranges from dark purple, 160m altitude, to yellow, 205m, for the first two rows and from -30m to 30m for the visualisation of $\Delta z$.}
    \label{fig:crop_visualisation}
 \end{figure*}

 
In Fig. \ref{fig:results_feature_mapping}, we show the \textit{IoU} (in $\%$) results between the different mapping methods: no feature mapping, RFF and SIREN.
In Fig. \ref{fig:auc} of the Appendix, we show the results for the AUC metric.
In Fig. \ref{fig:crop_visualisation}, we show some resulting crops trained with no feature mapping, RFF and SIREN.
We show additional crops as well as the whole map for data (3) in Fig.\ref{fig:whole_visualisation}, \ref{fig:crop1_visualisation}, \ref{fig:crop2_visualisation} and \ref{fig:subcrop2_visualisation}  of the Appendix.
For both metrics, \textit{IoU} and AUC, RFF outperforms SIREN and the default configuration by a fair margin. 
In terms of average performance (given in $\%$) with standard deviation, RFF reaches an \textit{IoU} of $53.0 \pm 12.$  and an AUC of $97.6 \pm 0.7$ and, SIREN $39.0 \pm 10.$ and  $95.9 \pm 2.8$, and not using any mapping reaches $30. \pm 15.$  and $95.9 \pm 3.8$.
A deconvolution of the reported values is given in Table~\ref{fig:full-results} in the Appendix.
From the visualisation in Fig. \ref{fig:crop_visualisation}, no feature mapping leads to a reconstruction where the building delimitation is unclear and fuzzy.
SIREN gives the sharpest reconstruction with close to no noise between the buildings.
Conversely, SIREN's projection onto the support of the second timestamp, given in the final row, is subject to many false positives along building boundaries.
The RFF method produces distinctive buildings, like SIREN, but with a noisier output.
However, the number of false positives in the final row is smaller.
This ablation study shows the necessity of feature mapping to achieve good \textit{IoU} and, therefore, a good reconstruction.
Capturing high-frequencies is essential to our current problem due to the verticality of buildings.

\subsection{Hyper-parameter influence}
\begin{figure*}[t]
    \centering
    \begin{subfigure}[b]{0.45\textwidth}
        \centering
        \includegraphics[width=0.9\linewidth, height=4.83cm]{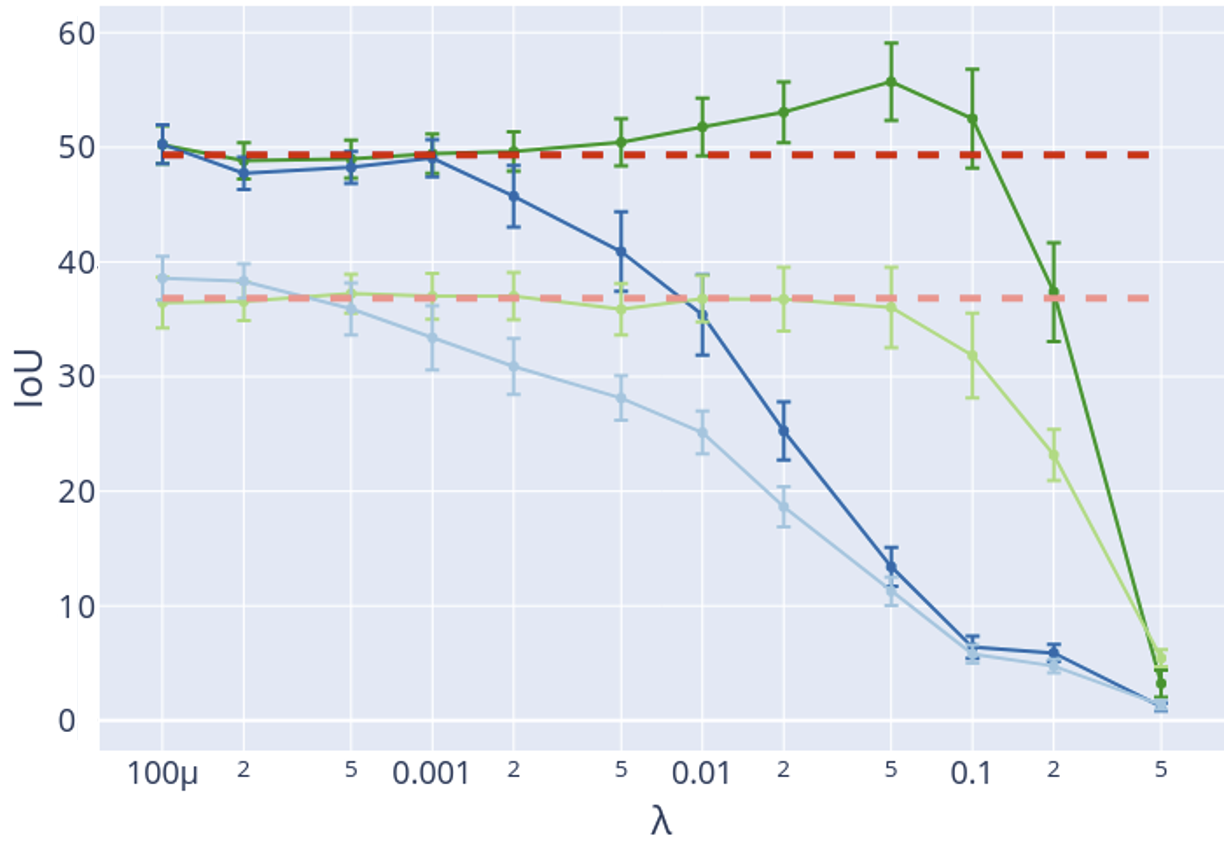}
        \caption{$IoU$ vs $\lambda$}
        \label{fig:iou_lambda}
    \end{subfigure}
    \hfill
    \begin{subfigure}[b]{0.54\textwidth}
        \centering
        \includegraphics[width=.9\linewidth, height=5.cm]{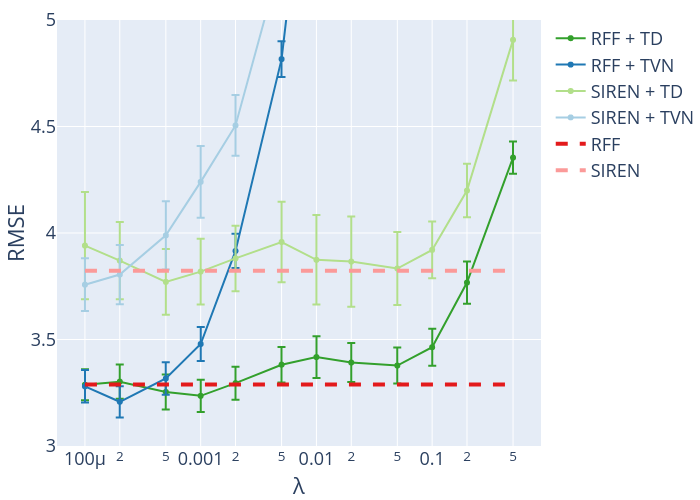}
        \caption{$RMSE$ vs $\lambda$}
        \label{fig:mse_lambda}
    \end{subfigure}
    \caption{Influence of parameter $\lambda$, refering to either $\lambda_{TD}$ or $\lambda_{TVN}$, with respect to the \textit{IoU} or reconstruction metric. We show $95\%$ confidence intervals for each estimator. Each points followed other hyper-parameter selection procedures (20 attemps to minimise the reconstruction) and was repeated 20 times. In red dashes, we show the model with $\lambda$ set to 0.}
    \label{fig:hyper}
 \end{figure*}

In Fig. \ref{fig:hyper}, we show a study on the regularisation parameter $\lambda_{TD}$ and $\lambda_\text{TVN}$ for both RFF, in darker and SIREN, in lighter colours.
We measure the \textit{IoU} and RMSE for both and compare them to the setting without penalty.
Naturally, a too-strong penalty damages the performance, and a too-low value will render the penalisation negligible.
Only for the method using RFF and TD regularisation do we see a $6\%$ improvement in \textit{IoU} compared to the baseline. 
This optimal $\lambda_{TD}$ does not necessarily correspond to a minimal RMSE, metric used for the validation scheme.
Similarly to Section \ref{sec:featuremap}, RFF features obtain better performance on both metrics and a more stable reconstruction noticeable by a lower RMSE and smaller confidence intervals.

The literature reports better reconstruction for SIREN over the RFF methods.
However, in our situation, SIREN gives lower performances in terms of \textit{IoU} and MSE.
SIREN suffers more from the mathematical formulation given in Section \ref{sec:method}, which is ill-posed because of many points on the sides of the buildings.
As SIREN induces qualitatively a better reconstruction, i.e. sharper edges and less noise. 
Having samples from the PC sharing similar geographical coordinates but radically different altitudes (along the verticality of the building) leads SIREN to 
slightly misplace the boundaries of the buildings, leading to many false positives along their edges, and hence a lower \textit{IoU}.
SIREN's MSE is higher, as the MSE penalises false positives more strongly as they correspond to larger differences between the ground truth and the prediction.
The sharper edges of SIREN, compared to RFF, are harmful in terms of \textit{IoU} and MSE.
In other words, SIREN's reconstruction is sharper and, due to noise, wrongly estimates the building size. 
In contrast, the RFF method has softer edges, i.e. less overfitting induced by capturing fewer high-frequencies. This leads to a better mean error along the building edges and, to a lower minimisation of the MSE and fewer false positives.

 \newcommand{\BEST}{\bfseries}

 \begin{table*}[!ht]
 \caption{Comparaison to state-of-the-art on the $IoU$ metric (in $\%$), we only report the best configuration when no feature mapping and SIREN. We show the best performing model in each row. D denotes the model with two DNN given by equation (\ref{eq:loss2}) and S the model with a single DNN given by equation (\ref{eq:loss1}). The complete table can be found in Table \ref{fig:full-results} of the Appendix. \label{tab:results}}\vspace{-.0in}
 \begin{tabular}{c|cc|c|c|cccccc}
 \toprule
 \multirow{2}{*}{Data} &  \multirow{2}{*}{M3C2 \cite{lague2013accurate}}    &    \multirow{2}{*}{OT \cite{fiorucci2023optimal}}          & None  & SIREN & \multicolumn{6}{c}{RFF (Proposed Method)}  \\ \cmidrule{4-11} 
   &  &  & S+TVN & S   & D & D+TVN & S & S+TVN     & S+TD    & S+TVN+TD \\ \toprule
 (1)  & 29.87  & 40.65         & 37.22 & 40.14 & 50.13 & 44.68 & 52.49 & {\BEST 55.87} & 54.73         & 49.68           \\
 (2)  & 53.73  & 55.20         & 45.22 & 53.98 & 57.39 & 59.33 & 56.45 & {\BEST 61.17} & 60.34         & 59.52           \\
 (3)  & 38.72  & 39.26         & 33.11 & 38.57 & 46.54 & 43.17 & 51.87 & 46.94         & {\BEST 54.00} & 53.33           \\
 (4)  & 35.01  & 39.89         & 33.54 & 39.01 & 48.62 & 49.70 & 51.38 & 51.10         & 53.40         & {\BEST 53.99}   \\
 (5)  & 37.78  & {\BEST 48.17} & 37.97 & 40.48 & 42.55 & 43.16 & 42.95 & 43.26         & 40.55         & 47.17           \\ \midrule
 Avg  & 39.02  & 44.63         & 37.41 & 42.43 & 49.04 & 48.00 & 51.02 & 51.67         & 52.60         & {\BEST 52.74}   \\ \bottomrule
 \end{tabular}
 \end{table*}
 
\subsection{Comparison to state of the art}
In Table \ref{tab:results}, we benchmark methods on the simulated dataset and compare the previous state-of-the-art in unsupervised detection, M3C2 \cite{lague2013accurate}, OT \cite{fiorucci2023optimal} and SIREN to our method. 
RFF outperforms the other methods by a large margin, about $13\%$ and $8\%$  in \textit{IoU} over the previous state of the art.
In addition, the previous state-of-the-art maximised the \textit{IoU} with respect to a predefined threshold, whereas our method is completely unsupervised.
For example, SIREN still improves over the previous state-of-the-art even if the results do not show this because the metric report for SIREN is unbiased.

The experiments favour the use of one single function for both timestamps.
We notice a difference of $2\%$ in \textit{IoU} between using a single model (S) and two models (D) when no regularisation is applied. 
Expectedly, the best-achieving model uses regularisation. 


The final performed comparison is with DC3DCD on Urb3DCD-v2, and the results are shown in Table \ref{tab:results2}.
To compare fairly, we use the same weakly supervised setting as the authors to map unlabelled classes, which for us corresponds to our binned distance, to labelled classes~\cite{de2023dc3dcd}.
In Table \ref{tab:results2}, we compare, DNN to DNN,  and show that INR outperforms the only other unsupervised DNN for CD on the building classes.
The best-achieving models on the first dataset were optimised for building change and, hence, performed poorly on the vegetation classes for the second version.
The model hyper-parameters and post-processing should be adapted, in particular relaxing the TVN penalty (which penalises the detection of small objects), the TD penalty (detection of small growth) and adapting the post-processing for the specific task of vegetation changes.
DC3DCD shows better performance than our proposed method when coupled with specific manual input features \cite{de2023dc3dcd}. 
However, it should be noted that our model comprises approx. 600K trainable parameters, whereas DC3DCD has more than 100M parameters \cite{thomas2019kpconv}. \footnote{This is only an approximation as the precise number requires the number of kernel points, which is unknown.}


\newcommand{\best}[1]{\mathbf{#1}}

\begin{table}[!ht]
\centering
\caption{Comparaison to state-of-the-art on the $IoU$ metric per class on Urb3DCD-v2 (in $\%$). In each column, we highlight the best achieving model.  \label{tab:results2}}
\resizebox{0.7\linewidth}{!}{
\begin{tabular}{c|c|c|c}
\toprule
Method    & DC3DCD & SIREN+S+TVN+TD &  RFF+S+TVN+TD \\ 
\midrule
Unchanged       & $\best{90.90 \pm 0.70}$   & $84.83 \pm 6.32$  & $87.47 \pm 3.17$ \\
New building    & $64.06 \pm 5.13$ & $62.62 \pm 11.4$ & $\best{71.81 \pm 2.76}$   \\
Demolition      & $54.35 \pm 3.84$ & $47.92 \pm 10.51$  & $\best{57.63 \pm 5.09}$  \\
New veg.        & $\best{58.14 \pm 20.03}$ & $4.26 \pm 4.87$   & $5.44 \pm 5.88$  \\
Veg. Growth     & $1.45 \pm 2.05$  & $0.62 \pm 0.29$  &   $\best{1.54 \pm 1.42}$   \\
Missing veg.    & $0.94 \pm 0.78$  &  $3.89 \pm 4.83$  & $\best{8.82 \pm 7.53}$  \\
Mobile Object   & $\best{47.57 \pm 2.58}$   & $0.26 \pm 0.19$      & $0.58 \pm 0.26$     \\
\bottomrule
\end{tabular}}
\vspace{-.0in}
\end{table}

\subsection{Application to Cultural heritage}
\begin{figure}
    \centering
    \includegraphics[width=0.7\linewidth]{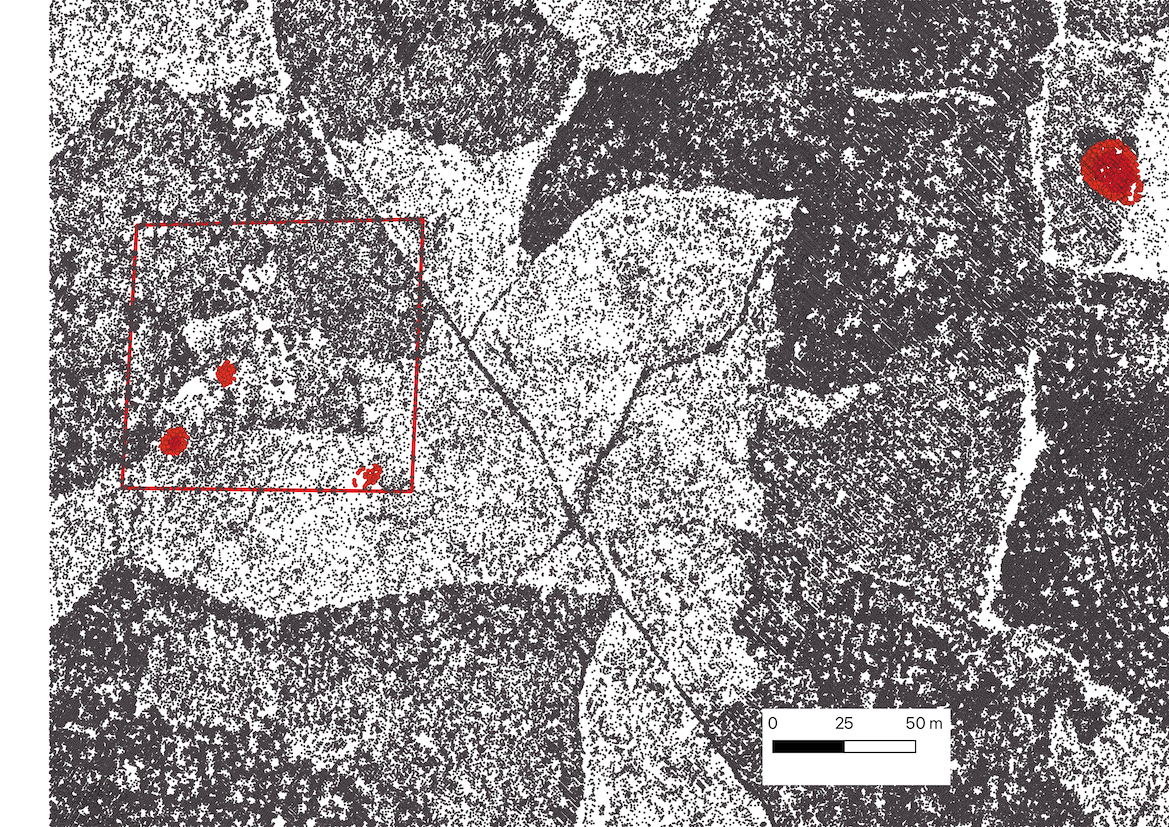}
    \caption{Detection of looting pits on a bi-temporal pair of LiDAR point clouds. The ground truth bounding box identified the geographical area where looting occurred. The red points represent detected looting pits.}
    \label{fig:looting}
 \end{figure}

The two-fold purpose of using LiDAR PCs to identify looting activities is to validate model `S+RFF+TVN+TD' on real (non-simulated) bi-temporal pairs of LiDAR PCs and assess its capability to detect looting, which is a pressing global-scale problem. 
We processed one bi-temporal pair of LiDAR PCs acquired over the Phnom Kulen region (Cambodia), where temples and ancient dams of the Angkor era are largely obscured by thick and closed canopies. 
Still, LiDAR deals well with this environment thanks to its capability to penetrate landscapes covered by continuous vegetation. 
The first, $pc_0$, was acquired in $2012$, and the second, $pc_1$, in $2015$. 
Fig.~\ref{fig:looting} shows detected looting.
Archaeologists drew the red bounding box to identify an area where looting occurred. 
The archaeologists verified the predicted changes through visual inspection and confirmed that all the looting pits inside the bounding box were correctly identified. 
The false positive in the top right part can be easily filtered due to the $20$ meter diameter, which is too big to be considered as a looting pit.

\section{Conclusion}
The amount of Earth observation acquired with 3D LiDAR data is rising exponentially, which opens up the possibility of monitoring human activity through CD algorithms.
In particular, we focused on urban planning and looting activities identification.
Thanks to advances in DNN, we can now estimate and reconstruct large areas with high precision that allows passing the reconstructed surfaces to downstream tasks.
However, the amount of training data and their discrete modelling limits their application to real scenarios.
To address these issues, we propose a novel unsupervised grid-agnostic scheme for CD based on surface reconstruction and clustering, which achieves $52.74\%$ \textit{IoU} (in $\%$), surpassing previous state-of-the-art M3C2 and OT by $10\%$ on average on the Urb3DCD dataset.
Moreover, we demonstrated in this paper that RRF mapping outperforms SIREN for CD on PC data acquired from airborne LiDAR sensors and allows us to identify looting activity correctly.


\vspace{-0.3cm}\section*{Code availability}
The code is made fully available at the following URL: \href{https://github.com/PeterJackNaylor/NN-4-change-detection/}{NN-4-change-detection}, it include the experimental pipelines as well as a colab notebook for training.
The code runs efficiently with hyperparameter selection thanks to the Optuna \cite{akiba2019optuna} package. We use GPU computation with PyTorch \cite{NEURIPS2019_9015} and combine each experiment into a pipeline with Nextflow \cite{di2017nextflow} for easy reproducibility.

\section*{Aknowledgement}
MY was supported by MEXT KAKENHI 20H04243 and partly supported by MEXT KAKENHI 21H04874. 
MF was supported by the European Union’s Horizon 2020 research and innovation programme under grant agreement No 101027956. 
We thank Damian Evans for providing us with a pair of LiDAR PCs for looting identification. 
We are also thankful for the RAIDEN computing system and its support team at the RIKEN AIP, which we used for our experiments.

\newcommand{\suptitle}{Supplementary Material: \\ Implicit neural representation for change detection}
\clearpage
{
\small
\bibliographystyle{unsrt}
\bibliography{egbib}
}
\clearpage


\renewcommand{\thesection}{A.\arabic{section}}
\renewcommand{\thetable}{A.\arabic{table}}
\renewcommand{\thefigure}{A.\arabic{figure}}
\setcounter{section}{0}
\setcounter{table}{0}
\setcounter{figure}{0}

\vbox{%
\hsize\textwidth
\linewidth\hsize
\vskip 0.1in
\rule{\linewidth}{2pt}
\centering
{\LARGE\sc \suptitle \par}
\rule{\linewidth}{2pt}

\vskip 0.1in
\def\And{%
  \end{tabular}\hfil\linebreak[0]\hfil%
  \begin{tabular}[t]{c}\bf\rule{\z@}{24\p@}\ignorespaces%
}
\def\AND{%
  \end{tabular}\hfil\linebreak[4]\hfil%
  \begin{tabular}[t]{c}\bf\rule{\z@}{24\p@}\ignorespaces%
}
\begin{tabular}[t]{c}\bf
Peter Naylor\\
RIKEN AIP\\
Kyoto, Japan\\
{\tt\small peter.naylor@riken.jp}
\and
Diego Di Carlo\\
RIKEN AIP\\
Kyoto, Japan\\
{\tt\small diego.dicarlo@riken.jp}
\and
Arianna Traviglia\\
Istituto Italiano di Tecnologia\\
Venice, Italy\\
{\tt\small arianna.traviglia@iit.it}
\and
Makoto Yamada\\
OIST\\
Okinawa, Japan\\
{\tt\small makoto.yamada@oist.jp}
\and
Marco Fiorucci\\
Istituto Italiano di Tecnologia\\
Venice, Italy\\
{\tt\small marco.fiorucci@iit.it}
\end{tabular}%
\vskip 0.4in \center{ }   \vskip 0.2in 
}

We show in the Supplementary Material results of all the methods applieds to each data configuration in Table~\ref{fig:full-results}.
We summarise all the possible architectures that we optimise from in Table~\ref{tab:model-recap} and show an example model in Fig.~\ref{fig:model}.
In addition, we give the AUC plot between RFF, SIREN and no feature mapping in Fig.~\ref{fig:auc}.
We show in Fig.~\ref{fig:diffz} the distribution of a typical $\Delta z$, mapped with the true labels and mapped with the predicted GMM labels.
Finally, we show an additional samples of surface reconstruction, namely,  the whole map in Fig.~\ref{fig:whole_visualisation}, a medium size crop in 
Fig.~\ref{fig:crop1_visualisation}, another medium-field size crop in Fig.~\ref{fig:crop2_visualisation} and a close field crop in Fig~\ref{fig:subcrop2_visualisation}.
In Fig.~\ref{fig:whole_visualisation}, we also show where each crops was extracted from, including the one in the main text.

\begin{table*}[htbp]
   \small
   \centering
   \subfloat[AUC for no feature mapping]{%
     \resizebox{0.47\linewidth}{!}{
      \begin{tabular}{c|cccccc}
         \toprule
         Data    &  D &  D+TVN &  S &  S+TD  &  S+TVN &  S+TD+TVN \\
         \midrule
         (1)  &  90.12 &      88.61 &  96.99 &       92.00  &      95.47 &           96.64 \\
         (2)  &  97.21 &      97.10 &  98.68 &       98.39  &      98.68 &           98.82  \\
         (3)  &  91.04 &      85.07 &  95.53 &       95.72  &      96.40 &           96.00 \\
         (4)  &  89.93 &      87.34 &  95.63 &       95.71  &      96.12 &           82.96 \\
         (5)  &  93.06 &      92.87 &  97.80 &       97.73  &      97.42 &           86.52\\
         \midrule
         Avg  &  92.27 &      90.20 &  96.92 &       95.91  &      96.82 &           92.19 \\
         \bottomrule
      \end{tabular}}
   }
   \subfloat[IoU for no feature mapping]{%
     \resizebox{0.47\linewidth}{!}{
     \begin{tabular}{c|cccccc}
      \toprule
      Data    &  D &  D+TVN &  S &  S+TD  &  S+TVN &  S+TD+TVN \\
      \midrule
      (1)  &  23.73 &      23.65 &  21.14 &       19.36  &      37.22 &           35.26 \\
      (2)  &  38.24 &      37.22 &  46.72 &       45.12  &      45.22 &           48.72 \\
      (3)  &  26.27 &      20.76 &  31.75 &       30.31  &      33.11 &           31.49 \\
      (4)  &  21.24 &      20.53 &  35.19 &       32.34  &      33.54 &           23.23 \\
      (5)  &  20.35 &      19.92 &  31.32 &       26.35  &      37.97 &           17.24 \\
       \midrule
      Avg  &  25.97 &      24.42 &  33.22 &       30.69  &      37.41 &           31.19\\
      \bottomrule
      \end{tabular}}
   }
   \hspace{0.5cm}%
   \subfloat[AUC for RFF]{%
     \resizebox{0.47\linewidth}{!}{
      \begin{tabular}{c|cccccc}
         \toprule
         Data    &  D &  D+TVN &  S &  S+TD &  S+TVN &  S+TD+TVN \\
         \midrule
         (1)  &  98.11 &      98.10 &  98.18 &       98.04  &      98.43 &           98.58 \\
         (2)  &  98.18 &      98.19 &  98.34 &       98.09  &      98.26 &           98.67 \\
         (3)  &  97.62 &      97.76 &  97.48 &       97.43  &      98.05 &           97.96 \\
         (4)  &  97.63 &      97.59 &  97.62 &       97.66  &      98.14 &           97.93 \\
         (5)  &  97.54 &      97.92 &  97.64 &       96.83  &      98.04 &           97.73 \\
         \midrule
         Avg  &  97.82 &      97.91 &  97.85 &       97.61  &      98.18 &           98.18 \\
         \bottomrule
         \end{tabular}}
   }
   \subfloat[IoU for RFF]{%
     \resizebox{0.47\linewidth}{!}{
      \begin{tabular}{c|cccccc}
         \toprule
         Data    &  D &  D+TVN &  S &  S+TD &  S+TVN &  S+TD+TVN \\
         \midrule
         (1)  &  50.12 &      44.67 &  52.49 &       54.73  &      55.86 &           49.67 \\
         (2)  &  57.38 &      59.32 &  56.45 &       60.33  &      61.16 &           59.52 \\
         (3)  &  46.54 &      43.16 &  51.87 &       53.99  &      46.94 &           53.33 \\
         (4)  &  48.62 &      49.69 &  51.38 &       53.40  &      51.10 &           53.99 \\
         (5)  &  42.55 &      43.15 &  42.95 &       40.55  &      43.25 &           47.17 \\
         \midrule
         Avg  &  49.04 &      48.00 &  51.03 &       52.60  &      51.66 &           52.74 \\
         \bottomrule
      \end{tabular}}
   }
   \hspace{0.5cm}%
   \subfloat[AUC for SIREN]{%
     \resizebox{0.47\linewidth}{!}{
      \begin{tabular}{c|cccccc}
         \toprule
         Data    &  D &  D+TVN &  S &  S+TD &  S+TVN &  S+TD+TVN \\
         \midrule
         (1)  &  97.80 &      97.45 &  97.74 &       97.49  &      98.04 &           97.77 \\
         (2)  &  98.13 &      97.79 &  98.44 &       98.24  &      98.30 &           97.93 \\
         (3)  &  97.02 &      97.45 &  97.55 &       96.41  &      96.27 &           96.60\\
         (4)  &  97.16 &      97.95 &  96.99 &       97.48 &      97.21 &           97.40 \\
         (5)  &  97.59 &      97.74 &  97.54 &       93.26 &      97.37 &           97.48 \\
         \midrule
         Avg  &  97.54 &      97.68 &  97.65 &       96.58 &      97.44 &           97.44 \\
         \bottomrule
         \end{tabular}}
   }
   \subfloat[IoU for SIREN]{%
     \resizebox{0.47\linewidth}{!}{
     \begin{tabular}{c|cccccc}
      \toprule
      Data    &  D &  D+TVN &  S &  S+TD &  S+TVN &  S+TD+TVN \\
      \midrule
      (1)  &  41.47 &      38.45 &  40.14 &       38.12 &      43.07 &           35.99 \\
      (2)  &  52.55 &      44.40 &  53.97 &       49.08 &      48.43 &           43.95 \\
      (3)  &  40.84 &      39.93 &  38.57 &       42.91 &      38.36 &           37.23 \\
      (4)  &  37.77 &      38.33 &  39.01 &       37.25 &      41.07 &           40.13 \\
      (5)  &  38.65 &      37.96 &  40.47 &       29.57 &      35.53 &           37.44 \\
      \midrule
      Avg  &  42.26 &      39.81 &  42.43 &       39.39 &      41.29 &           38.95 \\
      \bottomrule
      \end{tabular}}
   }
   \caption{Performance each data configuration for AUC and IoU (in $\%$). D denotes the model with two DNN given by equation (\textcolor{red}{1}) and S the model with a single DNN given by equation (\textcolor{red}{2}).}
   \label{fig:full-results}
 \end{table*}

\begin{table*}[!ht]
   \centering
   \resizebox{\textwidth}{!}{
   \begin{tabular}{c|c|c|c|c|c|c|c|c}
   \multicolumn{9}{c}{Models}                                                                                                                                            \\ \toprule
   default & default-BN  & default-L& skip-double           & skip-L-double         & skip-XL-double         & skip-ten             & skip-ten-only         & skip-twenty               \\ \midrule
   \multicolumn{9}{c}{Input $\gamma(\mathbf{v})$} \\ \midrule
   \multicolumn{2}{c|}{ } & FC-1024 &                      & FCS-1024 $(\times 2)$ & FCS-1024 $(\times 4$) & \multicolumn{3}{c}{ }                                           \\ \midrule
   \multicolumn{2}{c|}{ } & FC-512  & FCS-512 $(\times 2)$ & FCS-512 $(\times 2)$  & FCS-512 $(\times 4)$  & FCS-512 $(\times 10)$ & \multicolumn{2}{c}{ }                      \\ \midrule
   FC-256  & FC-256 + BN  & FC-256  & FCS-256 $(\times 2)$ & FCS-256 $(\times 2)$  & FCS-256 $(\times 3)$  & FCS-256               & FCS-256 $(\times 10)$ & FCS-256 $(\times 20)$ \\ \midrule
   FC-128  & FC-128 + BN  & FC-128  & FCS-128              & FCS-128 $(\times 2)$  & FCS-128 $(\times 2)$  & FCS-128               &                       & FCS-128 $(\times 2)$   \\ \midrule
   FC-64   & FC-64 + BN   & FC-64   & FCS-64               & FCS-64 $(\times 2)$   & FCS-64 $(\times 2)$   & FCS-64                &                       & FCS-64 $(\times 2)$    \\ \bottomrule 
   \multicolumn{9}{c}{Linear mapping to a 1 dimensional output} \\
   \bottomrule                                                                                               
   \end{tabular}
   }
\caption{Neural Network models. FC denotes fully connected layers with a given activation. BN-denotes batch normalisation. FCS denotes fully connected layers with a skip layer. For the transition when downsampling the dimension for fully connected skip layers, we add a simple FC that maps from one dimension to the other.}
\label{tab:model-recap}
\end{table*}

\begin{figure*}[!ht]
   \begin{center}
   \includegraphics[width=0.5\linewidth]{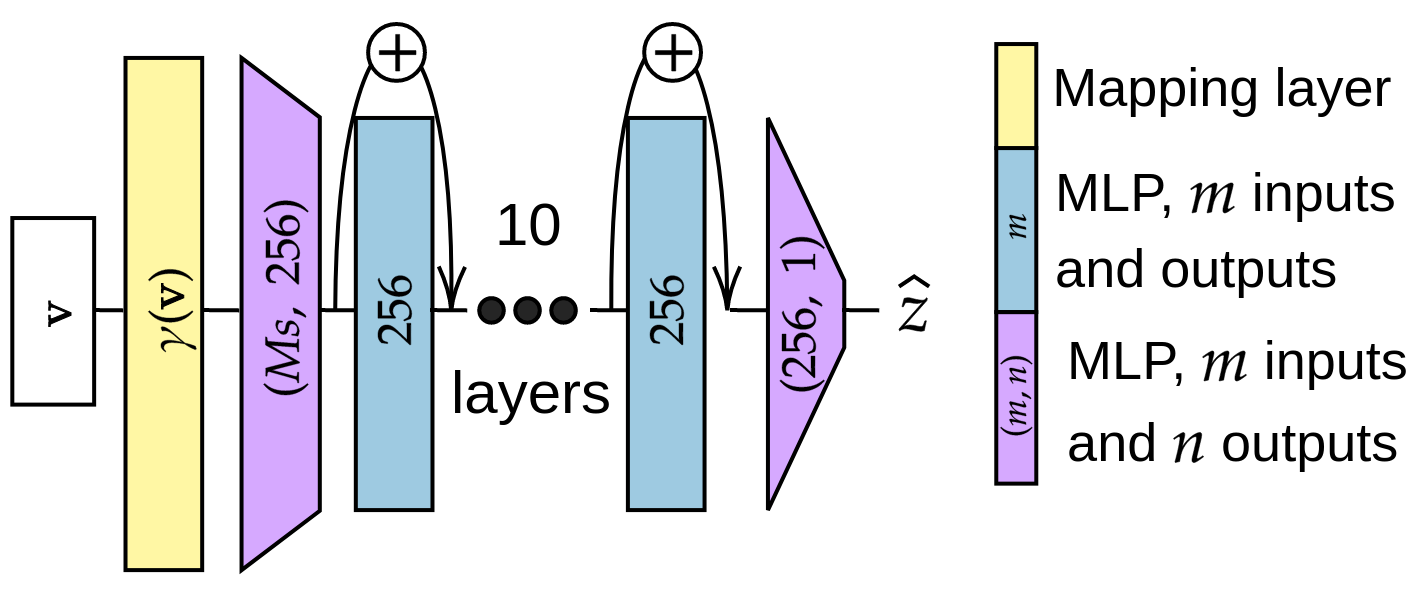}
   \end{center}
   \caption{Neural network model, `skip-ten-only'.}
   \label{fig:model}
\end{figure*}

\begin{figure*}[!ht]
   \centering
   \begin{subfigure}[b]{0.31\textwidth}
         \centering
         \includegraphics[width=\textwidth, height=4.cm]{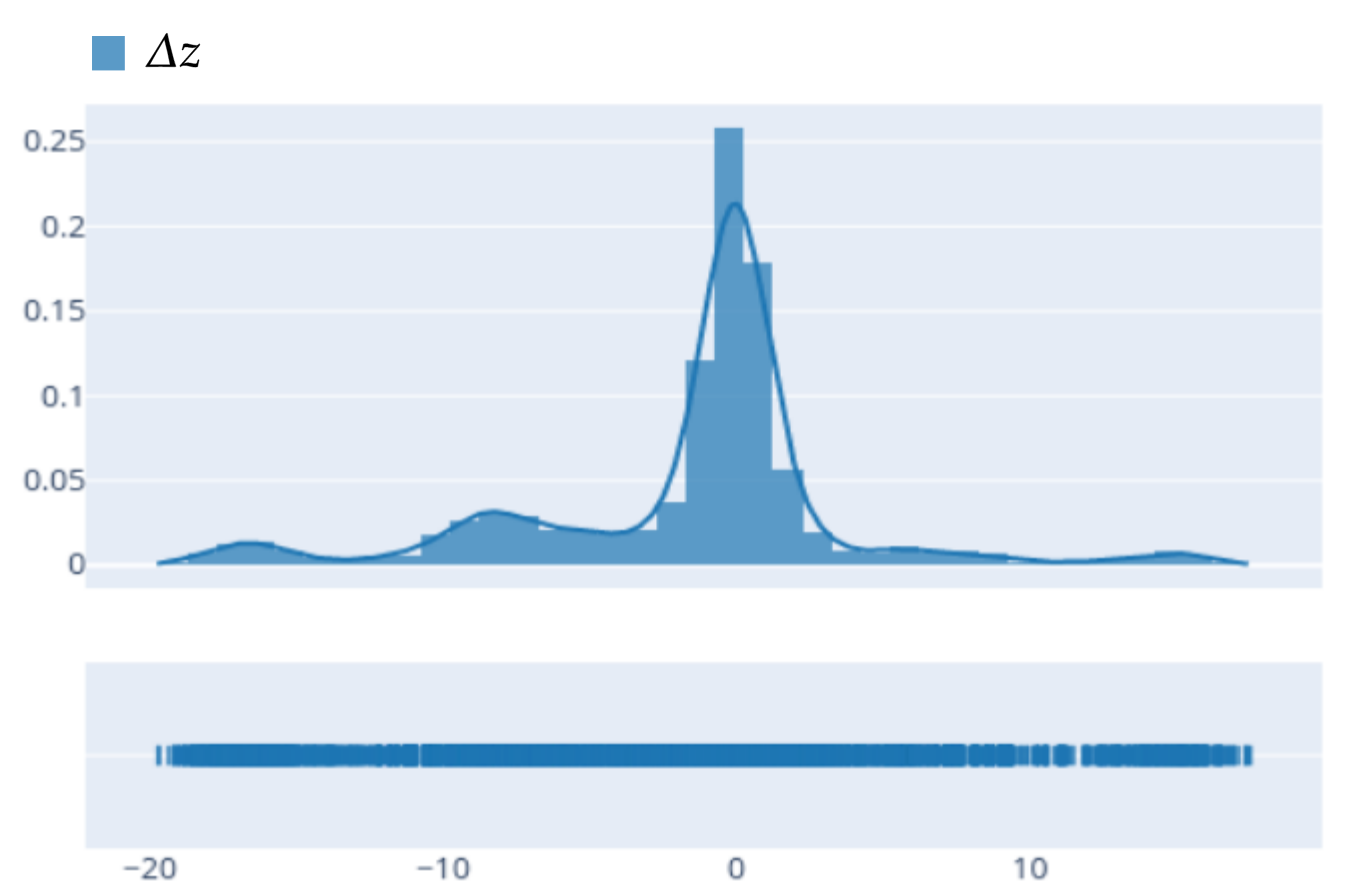}
         \caption{$\Delta z$}
         \label{fig:deltaz}
   \end{subfigure}
   \hfill
   \begin{subfigure}[b]{0.31\textwidth}
         \centering
         \includegraphics[width=\textwidth]{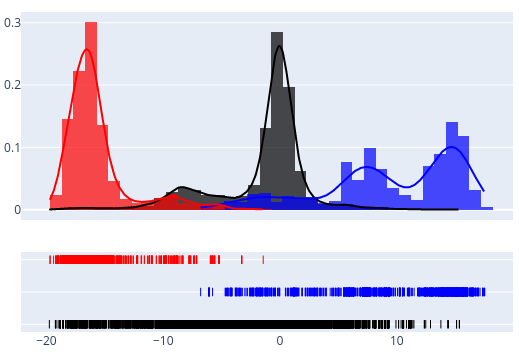}
         \caption{$\Delta z | class$}
         \label{fig:deltazlabel}
   \end{subfigure}
   \hfill
   \begin{subfigure}[b]{0.3\textwidth}
         \centering
         \includegraphics[width=\textwidth, height=4.05cm]{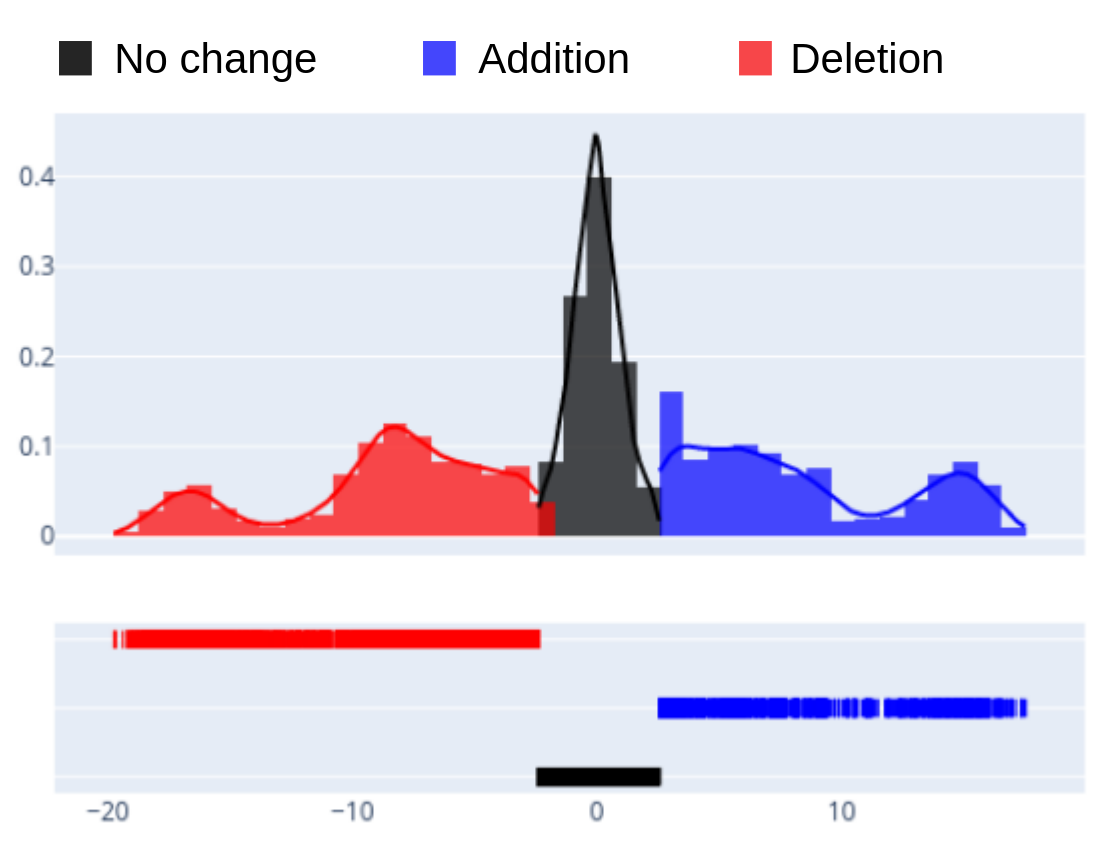}
         \caption{$\Delta z | \widehat{class}$}
         \label{fig:deltazpred}
   \end{subfigure}
      \caption{Distribution of $\Delta z$ and with predicted and true class label. This distribution was computed on a small subset with a single DNN (i.e. equation (\textcolor{red}{1}) with the RFF and no regularisation.}
      \label{fig:diffz}
\end{figure*}


\begin{figure*}[!ht]
   \centering
   \includegraphics[width=0.5\linewidth]{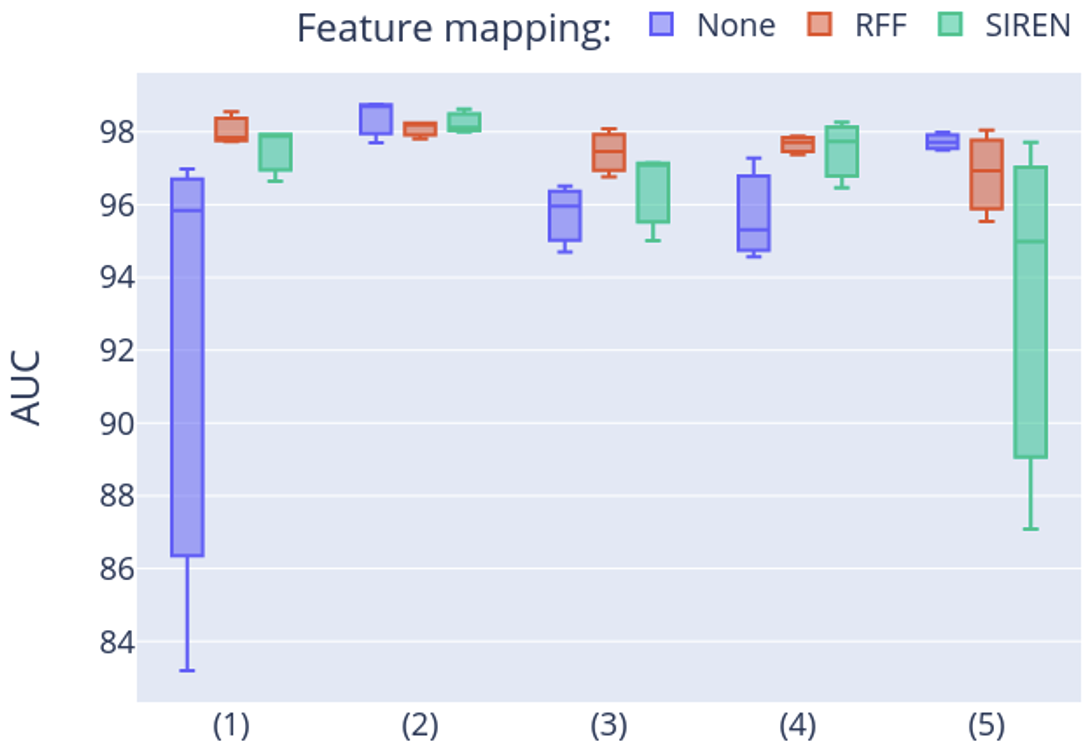}
   \caption{AUC results (in $\%$) for different feature mapping methods for every LiDAR airborn simulated dataset.}
   \label{fig:auc}
\end{figure*}

\begin{figure*}[!ht]
   \centering
   \includegraphics[width=1.\linewidth, height=18cm]{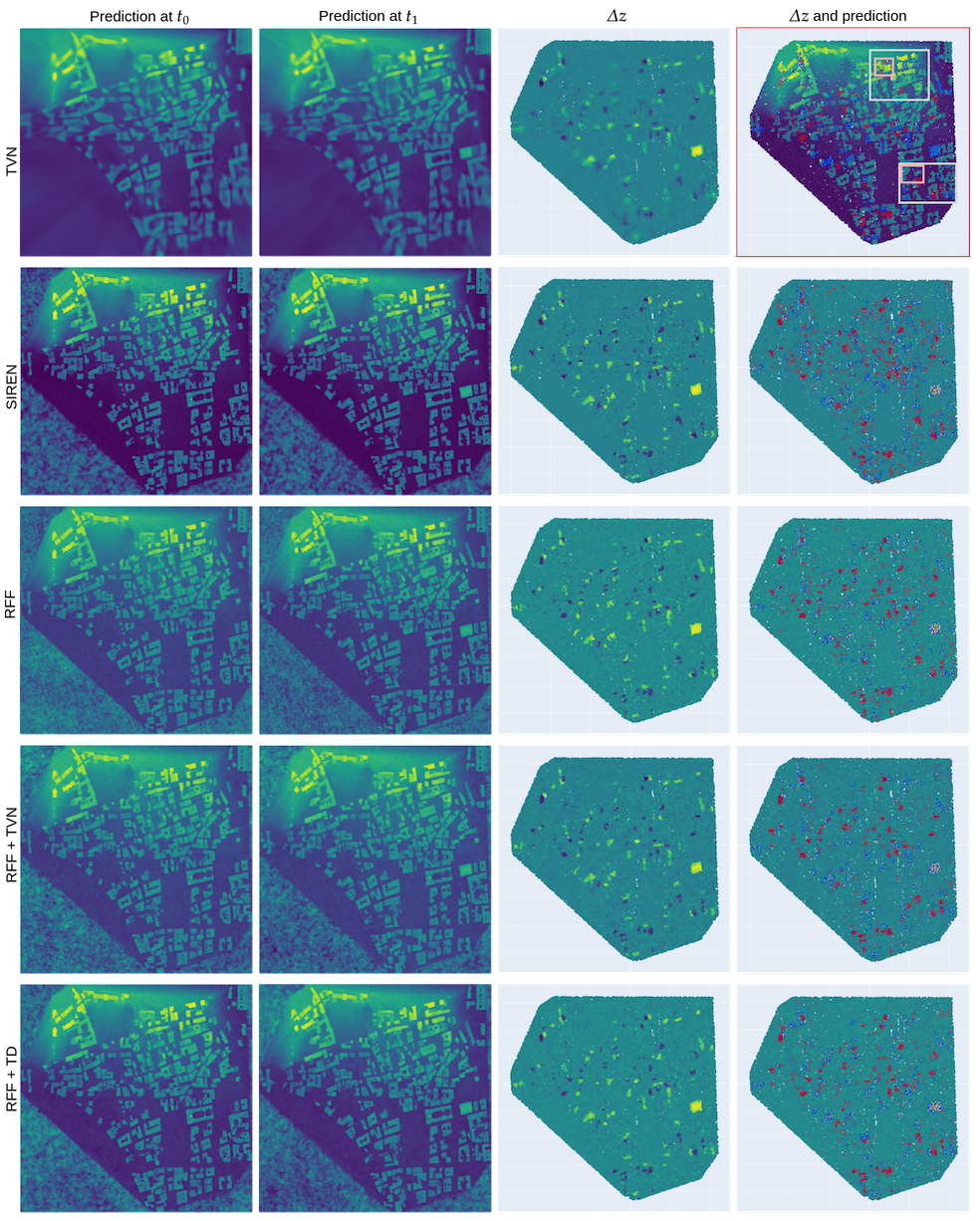}
   \caption{Visualisation of the whole map where in each row we show a different method comprising a single DNN. In the two first columns, we have the reconstruction of the surface along a regular grid for timestamp $t_0$ and $t_1$. In the third column, we show the difference $\Delta z$ on the support of $\mathcal{X}_1$ and in the fourth column we overlay these difference with the predicted labels from the GMM. Each column shows a different method. In the first row and in the last column we show the true cloud point overlaid with the ground truth. In addition, we show where the previous crops and sub-crops were extracted from in white and light red. To compair fairely, we range the color map from dark purple, 160m altitude, to yellow, 245m, for the first two rows and from -30m to 30m for the visualisation of $\Delta z$.}
   \label{fig:whole_visualisation}
\end{figure*}

\begin{figure*}[!ht]
   \centering
   \includegraphics[width=1.\linewidth]{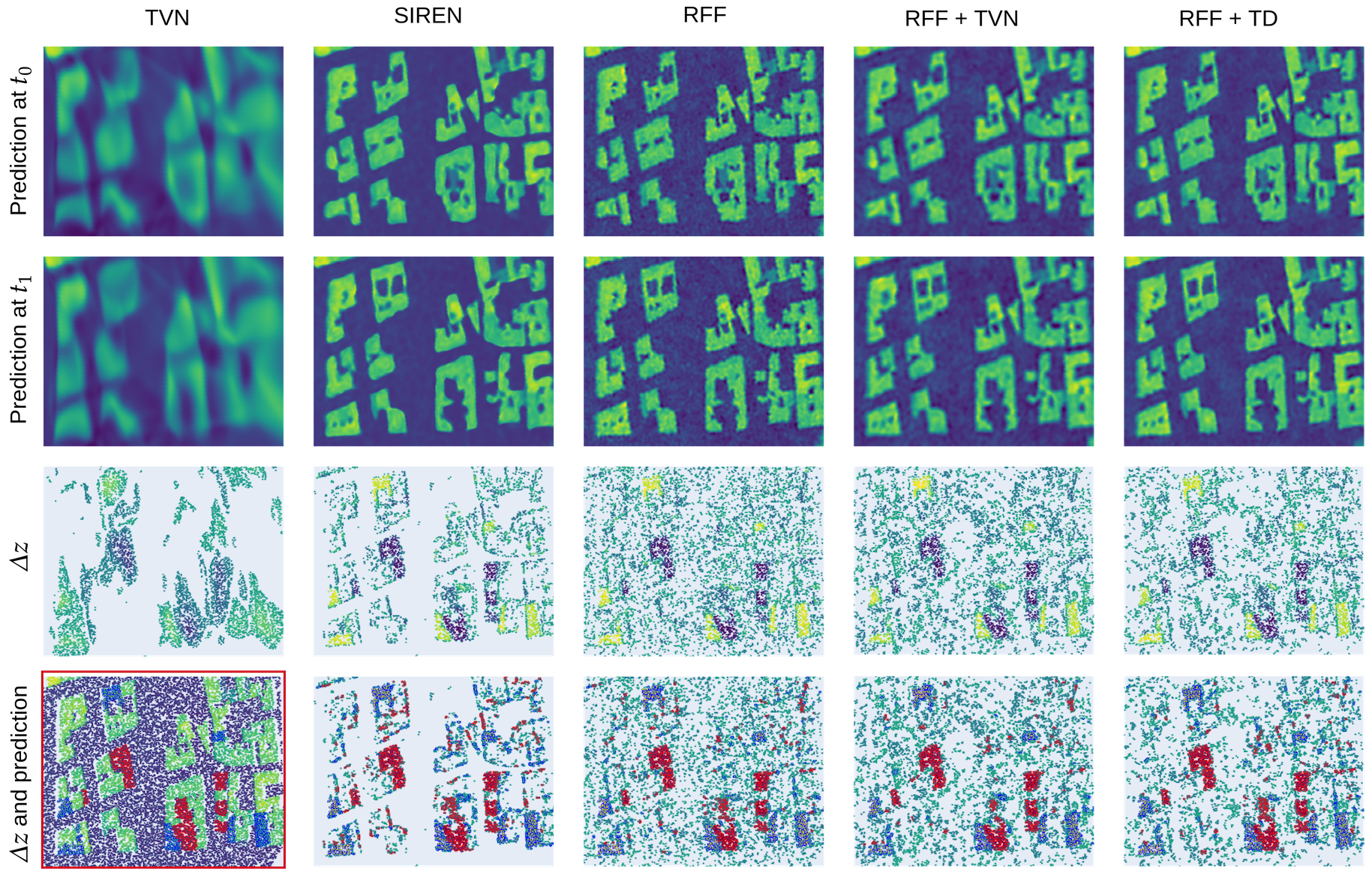}
   \caption{Visualisation of a crop number 1 where in each column we show a different method comprising a single DNN applied. In the two first rows, we have the reconstruction of the surface along a regular grid for timestamp $t_0$ and $t_1$. In the third row, we show the difference $\Delta z$ on the support of $\mathcal{X}_1$ and in the fourth row we overlay these difference with the predicted labels from the GMM, we filter out points where $|\Delta z| < 2$. Each column shows a different method. In the final row and in the first column we show the true cloud point overlaid with the ground truth. To compair fairely, we range the color map from dark purple, 160m altitude, to yellow, 245m, for the first two rows and from -30m to 30m for the visualisation of $\Delta z$.}
   \label{fig:crop1_visualisation}
\end{figure*}

\begin{figure*}[!ht]
   \centering
   \includegraphics[width=1.\linewidth]{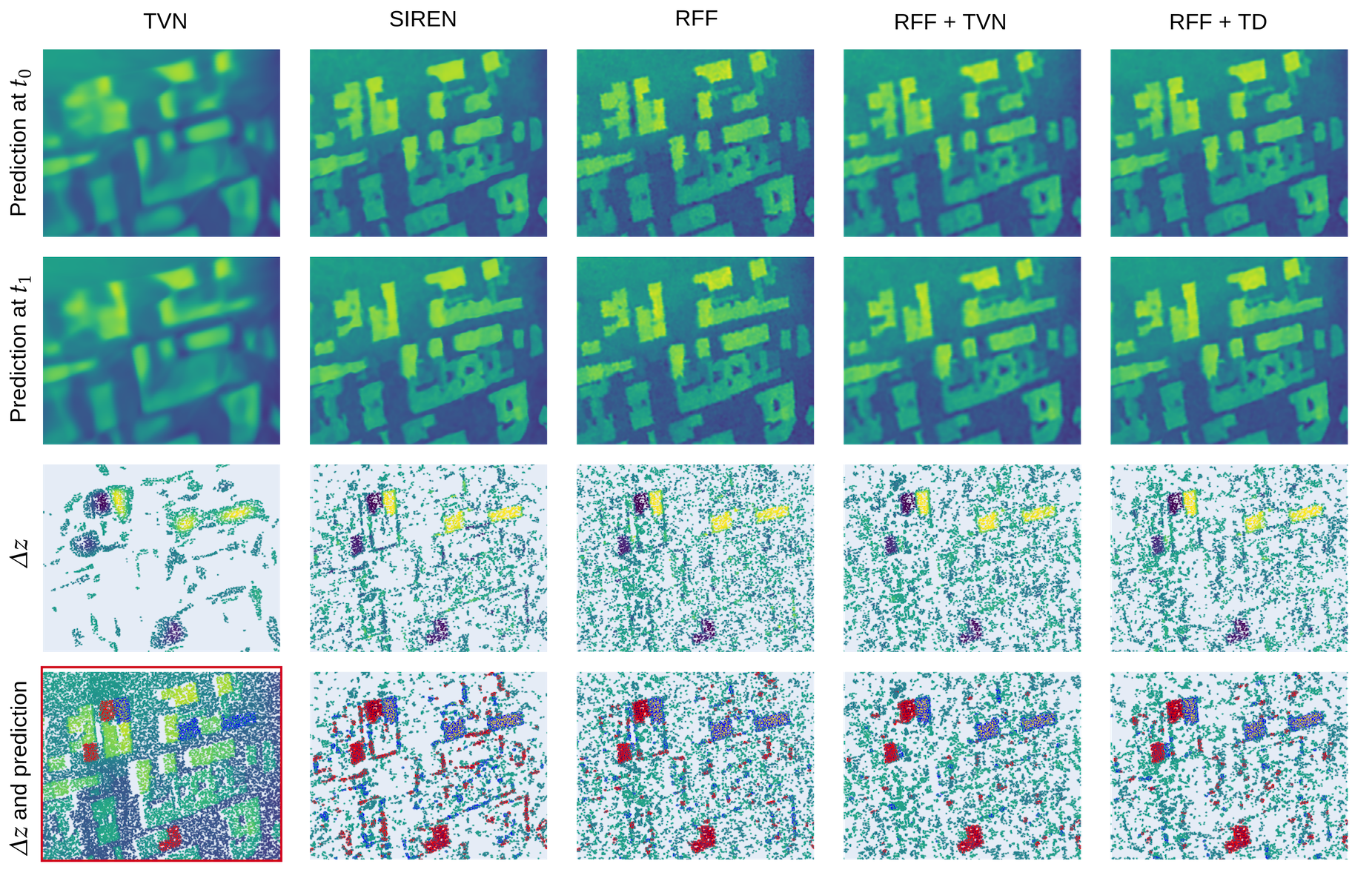}
   \caption{Visualisation of a crop number 2 where in each column we show a different method comprising a single DNN applied. In the two first rows, we have the reconstruction of the surface along a regular grid for timestamp $t_0$ and $t_1$. In the third row, we show the difference $\Delta z$ on the support of $\mathcal{X}_1$ and in the fourth row we overlay these difference with the predicted labels from the GMM, we filter out points where $|\Delta z| < 2$. Each column shows a different method. In the final row and in the first column we show the true cloud point overlaid with the ground truth. To compair fairely, we range the color map from dark purple, 160m altitude, to yellow, 245m, for the first two rows and from -30m to 30m for the visualisation of $\Delta z$.}
   \label{fig:crop2_visualisation}
\end{figure*}

\begin{figure*}[!ht]
   \centering
   \includegraphics[width=1.\linewidth]{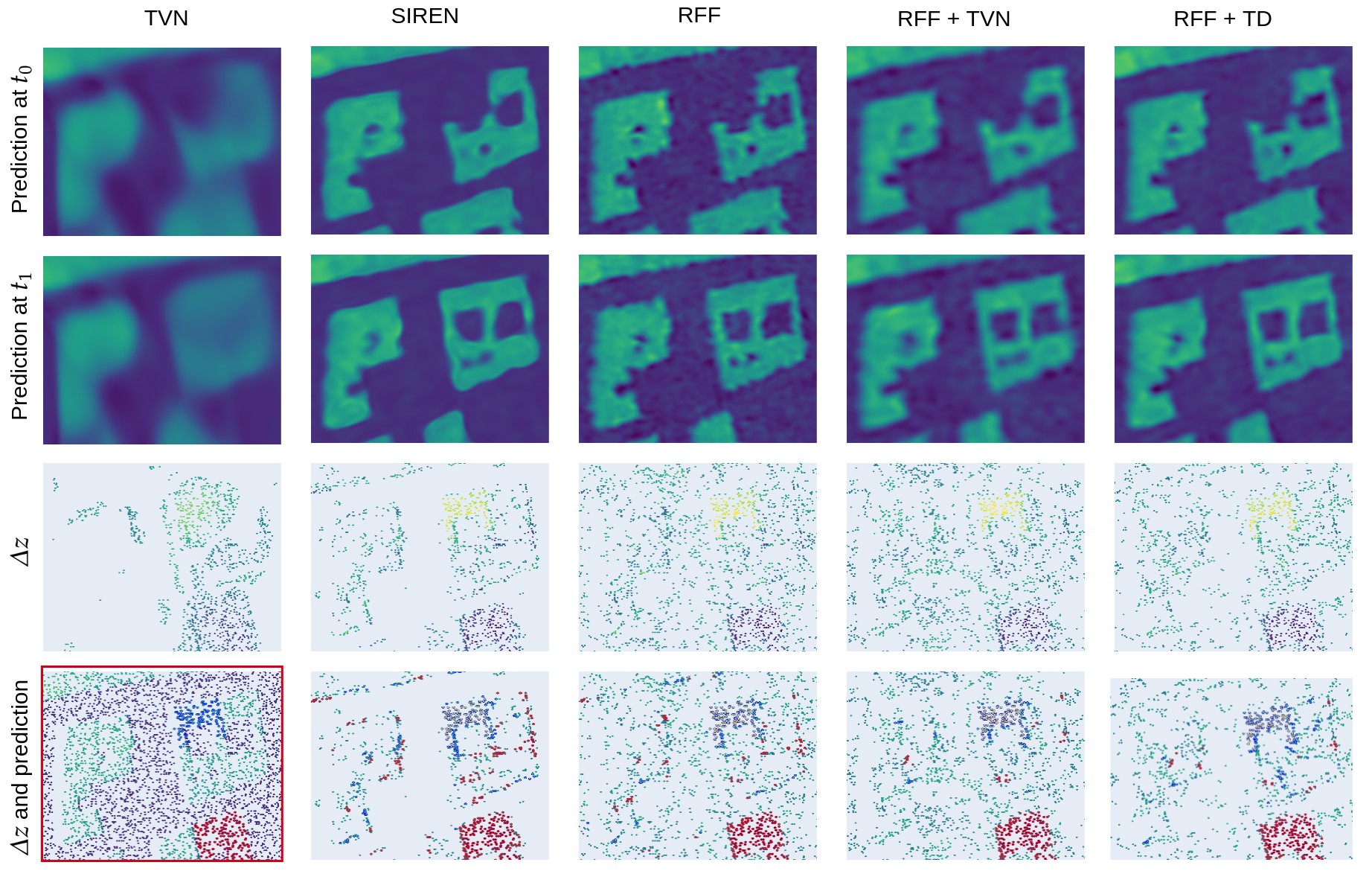}
   \caption{Visualisation of a sub-crop number 1 of crop number 1, where in each column we show a different method comprising a single DNN applied. In the two first rows, we have the reconstruction of the surface along a regular grid for timestamp $t_0$ and $t_1$. In the third row, we show the difference $\Delta z$ on the support of $\mathcal{X}_1$ and in the fourth row we overlay these difference with the predicted labels from the GMM, we filter out points where $|\Delta z| < 2$. Each column shows a different method. In the final row and in the first column we show the true cloud point overlaid with the ground truth. To compair fairely, we range the color map from dark purple, 160m altitude, to yellow, 245m, for the first two rows and from -30m to 30m for the visualisation of $\Delta z$.}
   \label{fig:subcrop2_visualisation}
\end{figure*}

\end{document}